\documentclass[runningheads]{llncs}
\usepackage{graphicx}

\usepackage{tikz}
\usepackage{comment}
\usepackage{amsmath,amssymb} 
\usepackage{color}

\usepackage[accsupp]{axessibility}  

\usepackage{graphicx}
\usepackage{amsmath}
\usepackage{amssymb}
\usepackage{booktabs}

\usepackage{multirow}
\usepackage{bm}
\usepackage{mathtools}

\newcommand{\wv}{{\mathbf w}}
\newcommand{\mv}{{\mathbf m}}

\newcommand{\zv}{{\mathbf h_T}}
\newcommand{\hv}{{\mathbf h_V}}

\begin{document}
\pagestyle{headings}
\mainmatter
\def\ECCVSubNumber{1897}  

\title{HiVLP: Hierarchical Vision-Language Pre-Training for Fast Image-Text Retrieval} 


\titlerunning{HiVLP}
%
\author{Feilong Chen  \and
Xiuyi Chen \and Jiaxin Shi \and\\  Duzhen Zhang \and Jianlong Chang \and Qi Tian }
\authorrunning{Chen et al.}
%
\institute{Huawei Cloud Computing \\
\email{ivess.chan@gmail.com, \{shijiaxin3, jianlong.chang, tian.qi1\}@huawei.com}}
\maketitle

\begin{abstract}
In the past few years, the emergence of vision-language pre-training (VLP) has brought cross-modal retrieval to a new era.
However, due to the latency and computation demand, it is commonly challenging to apply VLP in a real-time online retrieval system. 
To alleviate the defect, this paper proposes a \textbf{Hi}erarchical \textbf{V}ision-\textbf{}Language \textbf{P}re-Training (\textbf{HiVLP}) for fast Image-Text Retrieval (ITR). 
Specifically, we design a novel hierarchical retrieval objective, which uses the representation of different dimensions for coarse-to-fine ITR, i.e., using low-dimensional representation for large-scale coarse retrieval and high-dimensional representation for small-scale fine retrieval.
We evaluate our proposed HiVLP on two popular image-text retrieval benchmarks, i.e., Flickr30k and COCO. Extensive experiments demonstrate that our HiVLP not only has fast inference speed but also can be easily scaled to large-scale ITR scenarios. The detailed results show that HiVLP is $1,427$$\sim$$120,649\times$ faster than the fusion-based model UNITER and 2$\sim$5× faster than the fastest embedding-based model LightingDot in different candidate scenarios. It also achieves about +4.9 AR on COCO and +3.8 AR on Flickr30K than LightingDot and achieves comparable performance with the state-of-the-art (SOTA) fusion-based model METER.

\keywords{Image-Text Retrieval, Hierarchical Pre-training, Vision-Language Pre-training}
\end{abstract}

\section{Introduction}
\label{sec:intro}

\begin{figure}
    \centering
    \includegraphics[width=0.9\linewidth]{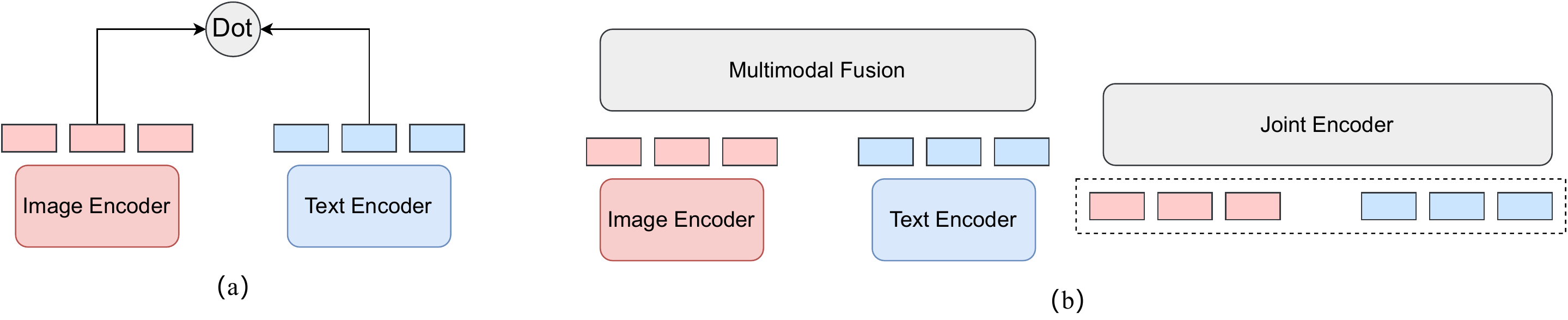}
    \caption{Two kinds of Image-text Retrieval models. (a) Embedding-based Models without cross-modal attention and fusion, (b) Fusion-based Models with cross-modal attention and fusion.}
    \label{fig:intro}
\end{figure}


With the explosive growth of multimodal data on the Internet, how to retrieve the expected data of different modalities fast and well is becoming an urgent problem. Image-text retrieval (ITR), which aims to retrieve the relevant images and texts across each other, has been widely studied because of its broad application prospects. Various benchmarks have been constructed to promote the development of ITR, such as Flickr30k~\cite{plummer2015flickr30k}, COCO~\cite{MSCOCO} and Multi30K~\cite{W16-3210}

In terms of the cross-modal interaction, existing methods of ITR can be divided into two categories: embedding-based models and fusion-based models. Embedding-based models~\cite{faghri2017vse++,wang2018learning,sun2021lightningdot} employ separate image encoder and text encoder to represent the image and text simultaneously and utilize their dot product for similarity matching as shown in Figure~\ref{fig:intro} (a). 
Fusion-based models~\cite{lee2018stacked,lee2019learning,wang2019camp,zhang2020context,lu2019vilbert,chen2020uniter} apply cross-modal attention between textual features and visual features for multimodal fusion as shown in Figure~\ref{fig:intro} (b). Some work~\cite{lee2018stacked,lee2019learning,wang2019camp} improve visual representation by utilizing advanced region-based visual encoder such as a pre-trained Faster R-CNN~\cite{RenHGS15}. 

With the recent remarkable success of pre-training technologies~\cite{devlin2019bert,dosovitskiy2020image}, the pretraining-finetune paradigm is widely demonstrated in vision-language areas~\cite{su2019vl,li2019visualbert,shen2021much,cho2021vlt5,tan2019lxmert,liu2021opt}, which also promotes the performance of ITR to new state-of-the-arts.
Different from early models~\cite{kiros2014unifying,faghri2017vse++,wang2018learning} that are train from scratch with ranking loss, current approaches~\cite{sun2021lightningdot,li2020oscar} pre-train the transformer-based~\cite{vaswani2017attention} image/text/joint encoders~\cite{devlin2019bert} with millions of weakly-supervised image-text pairs, and then finetune them with specific image-text downstream benchmarks.
The pre-training facilitates both embedding-based models~\cite{sun2021lightningdot} and fusion-based ones~\cite{lu2019vilbert,chen2020uniter,li2020oscar,gan2020large} and thus has become a standard technology in the ITR area.

Embedding-based models and fusion-based models have different advantages. From the accuracy aspect, fusion-based models inherently perform better than embedding-based models because of the adequate cross-modal interaction~\cite{anderson2018bottom}. However, fusion-based models have to fuse the query and all candidates online, which will cause an intolerable latency in realistic search scenarios.
Instead, embedding-based models enjoy the searching efficiency because these models disentangle the processing of images and texts and encode the massive candidates offline, which greatly accelerates the real-time searching~\cite{sun2021lightningdot}.
However, considering the huge amount of candidates on the Internet, it is still unaffordable to compute so many dot products for each query. For example, given $1e9$ candidates and $768$ hidden dimensions, the retrieval needs about $1.5$ trillion operations (multiplication plus addition), which is too costly even with a distributed cluster.
If we increase the feature dimension for better capability (\textit{i.e.}, $2048$ in \cite{liu2020fastbert}), the computation overhead will become heavier.


To this end, we propose a novel embedding-based model, HiVLP, 
short for \textbf{Hi}erarchical \textbf{V}ision-\textbf{}Language \textbf{P}re-Training, 
to enable fast and accurate real-time ITR over massive candidates.
We encode the query and candidates into hierarchical features of different dimensions 
to obtain the coarse-to-fine representation.
Low-dimensional features are used for fast and coarse retrieval over the large pool to narrow the candidate scope, and then high-dimensional features are used for accurate retrieval over the resultant small pool. 
Besides, hierarchical features are produced gradually by different encoder layers so that we can parallel the encoding process with the retrieval process, to reduce the overall latency further.

Specifically, we 
insert
an Early Output Layer (EOL) into the intermediate transformer layers, which converts the internal hidden states to external representation features. Starting from the bottom of the transformer stack, we increase the output dimension of EOL layer-by-layer to support a coarse-to-fine representation capability. 
For model training, we propose three training objectives to jointly train the visual transformer encoder and the linguistic transformer encoder: (1) Linguistic Representation Modeling (LRM) to ensure the model learns rich representations for both images and text, (2) Hierarchical Retrieval Learning (HRL) to ensure cross-modal alignment of different granularities, and (3) Vision-Language Matching (VLM) to obtain accurate matching.

We evaluate the proposed approach on two popular ITR benchmarks, and experiments show that our approach HiVLP is 1,427/6,426 $\times$ faster than the existing fusion-based model UNITER on Flick30k/COCO and achieves a higher performance. Moreover, we evaluate our approach on a larger candidate pool, HiVLP further shows its efficiency. 

Our contributions are summarized as follows:
\begin{itemize}
  \item We propose a novel approach HiVLP which utilizes hierarchical retrieval to accelerate image-text retrieval with slight performance degradation.
  \item We conduct extensive experiments on two popular ITR benchmarks and evaluate our approach under larger candidates' ITR settings. All the experiments show that our approach dramatically improves the speed of image-text retrieval and makes real-time ITR possible with low latency and high accuracy. HiVLP achieves with comparable accuracy and higher speed than both fusion-based models and embedding-based models. ($1,427$$\sim$$120,649\times$ faster than UNITER, $2$$\sim$$5\times$ faster than fastest embedding-based model LightingDot and comparable performance with METER). 
\end{itemize}

\section{Related Work}
Image text retrieval aims to retrieve images with text or text with images. The researchers' exploration of this task is divided into two aspects: (1) improving retrieval performance (2) accelerating retrieval speed. 

\subsection{Improving retrieval performance}
One way to improve retrieval performance is to apply heavy cross-modal interaction between vision and language. SCAN~\cite{lee2018stacked} utilize stacked cross-modal transformer blocks to handle the interaction. CAMP~\cite{wang2019camp} proposes cross-modal adaptive message passing to improve retrieval performance, which adaptively controls the information flow for message passing across modalities. CAAN~\cite{zhang2020context} obtains high retrieval performance via a unified context-aware attention network, which selectively focuses on critical local fragments (regions and words) by aggregating the global context. Another way to improve retrieval performance is to utilize pre-training technology~\cite{vlmo,wang2021simvlm}. Due to the success of BERT~\cite{devlin2019bert} in NLP and ViT~\cite{dosovitskiy2020image} in CV, more and more researchers begin to explore the visual-language pre-training models~\cite{su2019vl,li2019visualbert,shen2021much,cho2021vlt5,tan2019lxmert,liu2021opt,sun2021lightningdot}. Visual-language pre-training models learn the general vision-language representation and achieve amazing performance via finetuning themselves on image-text retrieval datasets.
  
\subsection{Accelerating retrieval speed}
There are some previous work~\cite{sun2021lightningdot,tu2021hashing,jia2021scaling} that aims to accelerate retrieval speed. LightingDot~\cite{sun2021lightningdot} utilizes two independent encoders to represent images and text and align them via dot product without cross-attention. Note that LightingDot uses a faster RCNN pre-trained to extract images. Due to the slow feature extraction speed of fast RCNN and the non-end-to-end training of LightingDot, this may lead to the reduction of speed and accuracy. ALIGN~\cite{jia2021scaling} leverages two transformers to learn visual and language representations via large-scale image-text pairs (1.8B). HEI~\cite{tu2021hashing} learns to map the original data points into short binary hash codes and coarsely preserve the heterologous matching relationship in order to accelerate retrieval speed.

\subsection{Hierarchical Retrieval}
SHAN~\cite{ji2021step} proposes a step-wise hierarchical alignment network to learn the alignment between images and text, which seems similar to our work. But in fact, the difference between our proposed approach and SHAN is very obvious. There are two main differences. First, SHAN learns hierarchical alignment after the image and text are encoded by encoders, but our proposed method is to let the model learn the alignment of different dimensional hierarchical features in the process of encoding the image and text by encoders. Second, the purpose of hierarchical retrieval is different. Hierarchical retrieval in SHAN is for full multimodal interaction, and ours is to speed up retrieval.


\begin{figure*}
    \centering
    \includegraphics[width=0.95\textwidth]{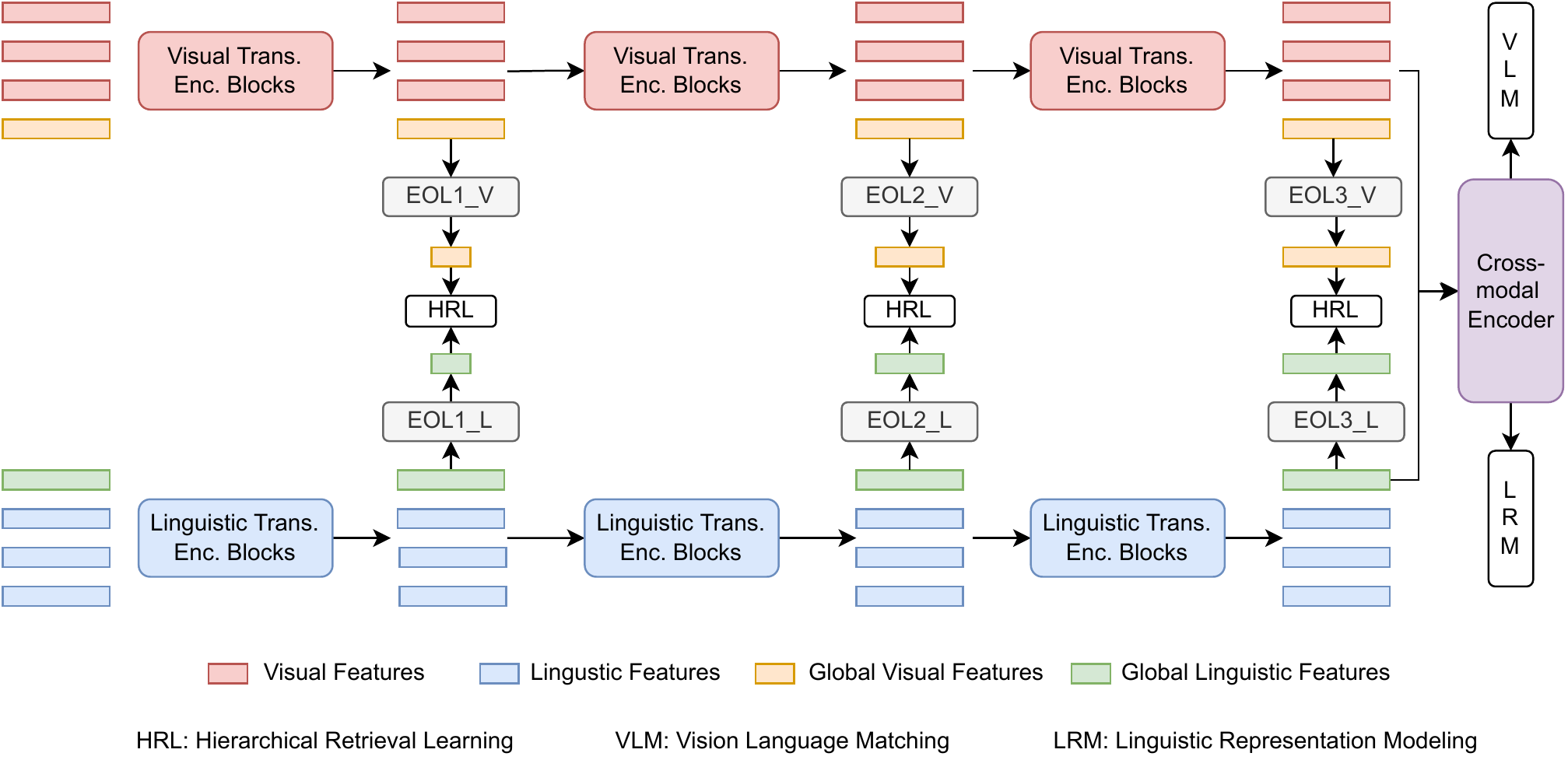}
    \caption{An overview of our proposed approach HiVLP. HiVLP inverts an Early Output Layer (EOL) into the intermediate transformer layers, which converts the internal hidden states to external representation features.}
    \label{fig:model}
\end{figure*}

\section{Approach}
In this section, we first introduce image-text retrieval and then propose our model with hierarchical pre-training, which consists of the visual transformer encoder, the linguistic transformer encoder and the cross-modal encoder as shown in Figure~\ref{fig:model}. Third, we present the objectives to train our model and finally, we describe the inference process by utilizing our hierarchical pre-training model.
\subsection{Task Definition}
Image-text retrieval can be divided into two sub-tasks: (1) image annotation for a given image and (2) image search for a given text. All the two sub-tasks are to rank the candidates (texts or images) for a given image or text.

\subsection{Hierarchical Pre-training Model}
Following BERT~\cite{devlin2019bert} and ViT~\cite{dosovitskiy2020image},  we utilize the transformer encoder modified from \cite{vaswani2017attention} as our visual encoder and linguistic encoder.

\subsubsection{Visual Transformer Encoder} For a given image from a image-text pair $(I, T)$, we reshape the 2D image $I \in \mathbb{R}^{H \times W \times C}$ into a sequence of 2D patches $x_p \in \mathbb{R}^{N^V \times (P^2 \cdot C)}$, where $(H, W)$ is the resolution of the image, $C$ is the channel number, $(P,P)$ is the resolution of each image patch, and $N^V=HW/P^2$ is the resulting number of patches. And then, we flatten the patches and map them to $d_h$ dimensions with a trainable linear projection. We regard the output of this projection as the patch embeddings.
We add a special token \texttt{[IMG]} to obtain the global representation of the input image. We use standard learnable 1D position embeddings to retain positional information. We add position embeddings to the patch embeddings to obtain visual input features $\bf x_V$. We encode $\bf x_V$ into multiple levels of visual representations $\hv = \{ \hv_0, \dots, \hv_{N^V} \}$ $(\hv_j \in \mathbb{R}^{d})$ by using $L^V$-stacked Transformer blocks, where the $l$-th Transformer block is denoted as $\mathbf{H}^l=\text{Transformer}(\mathbf{H}^{l-1}), l\in[1, L^V]$. Specifically, the representations $\mathbf{H}^l$ is calculated by using the multi-head self-attention~\cite{vaswani2017attention} as follows:  
\begin{equation}
    \mathbf{Q}=\mathbf{H}^{l-1}\mathbf{W}_l^{Q}, \mathbf{K}=\mathbf{H}^{l-1}\mathbf{W}_l^{K}, \mathbf{V}=\mathbf{H}^{l-1}\mathbf{W}_l^{V},
\end{equation}
\begin{equation}
    \mathbf{A}_l=\text{softmax}(\frac{\mathbf{Q}\mathbf{K}^T}{\sqrt{d_k}})\mathbf{V},
\end{equation}
where $\mathbf{W}_l^{Q}, \mathbf{W}_l^{K}, \mathbf{W}_l^{V}\in\mathbb{R}^{d_h\times d_k}$ are learnable weights for computing the queries, keys, and values respectively. Then $\mathbf{A}_l$ is passed into a feedforward layer to compute $\mathbf{H}^l$ for the next layer:
\begin{equation}\label{eq:mask}
    \mathbf{H}^l = {\rm FFN}(\mathbf{A}_l)
\end{equation}
We utilize $\mathbf{H}^l$ to represent $\hv^l$. We use $\hv_0^l$ to denote $l$-th global visual representation.

\paragraph{Linguistic Transformer Encoder}  For a given text from a image-text pair $(I, T)$, we employ WordPiece tokenizer~\cite{wu2016google} to split it into a word sequence $\mathbf{w}$, where each word is embedded with an absolute positional code following~\cite{devlin2019bert}, to obtain high-dimensional feature vectors $\wv = \{ \wv_0, \wv_1, ..., \wv_{N^T} \}$ ($\wv_j \in \mathbb{R}^{d_w}$, $N^T$ is the number of tokens). Similarly, we add a special token \texttt{[TXT]} and use $L^T$-stacked Transformer blocks to obtain the global representation  $\zv = \{ \zv_0, \dots, \zv_{N^T} \}$ $(\zv_j \in \mathbb{R}^{d})$ of linguistic features. We use $\zv_0^l$ to denote $l$-th global linguistic representation. 

\paragraph{Cross-modal Encoder} To obtain the fused representation of vision and language, our cross-modal encoder is applied a cross-attention~\cite{vaswani2017attention} and takes the representation from the linguistic transformer encoder as query input and the representation from the visual transformer encoder as key and value inputs. We regard the fused representation of \texttt{[TXT]} as multimodal representation.

\subsection{Training Objectives}
We utilize three objectives to train our model: (1) Linguistic Representation Modeling (LRM) to ensure the model learns rich representations for both images and text, (2) Hierarchical Retrieval Learning (HRL) to ensure hierarchical retrieval from rough retrieval to fine retrieval, (3) Vision-Language Matching (VLM) to obtain accurate matching. 


\subsubsection{Linguistic Representation Modeling (LRM)}
Similar to \cite{devlin2019bert,sun2021lightningdot}, we mask 15$\%$ tokens and train the model to reconstruct the masked tokens. Formally, we denote  $\mathbf{w}_\mv = \{ \mathbf{w}_{\mv_{1}}, \dots, \mathbf{w}_{\mv_M} \}$ as masked regions, where $\mathbf{m}\in \mathbb{N}^M$ is the set of masked indices of size $M$, randomly sampled from a natural number $\mathbb{N}$.  $\mathbf{w}_{\setminus \mathbf{m}}$ are the unmasked tokens. Linguistic Representation Modeling is supervised by:
\begin{equation}
\label{eq:imlm}
\begin{split}
    \mathcal{L}_{\text{LRM}}(T, I) = \mathbb{E}_{(I,T)\sim D}\mathrm{CE}(P_{\theta}(\wv_\mv &| \wv_{\setminus \mv},\, I)) \\
     = - \frac{1}{M} \sum_{k=1}^M \log P_{\theta_{\text{mlm}}}(\wv_{\mv_k} &| \zv_{\mv_k} + \hv_0^l)\,,
\end{split}
\end{equation}
where $\theta_{mlm}$ and the word probabilities $P_\theta$ are conditioned on the corresponding image $I$ via the global image representation $\hv_0^l$. Notice that we only utilize the final linguistic representation of the last layer to train the model with the linguistic representation modeling objective.

\begin{figure*}[t]
    \centering
    \includegraphics[width=0.9\textwidth]{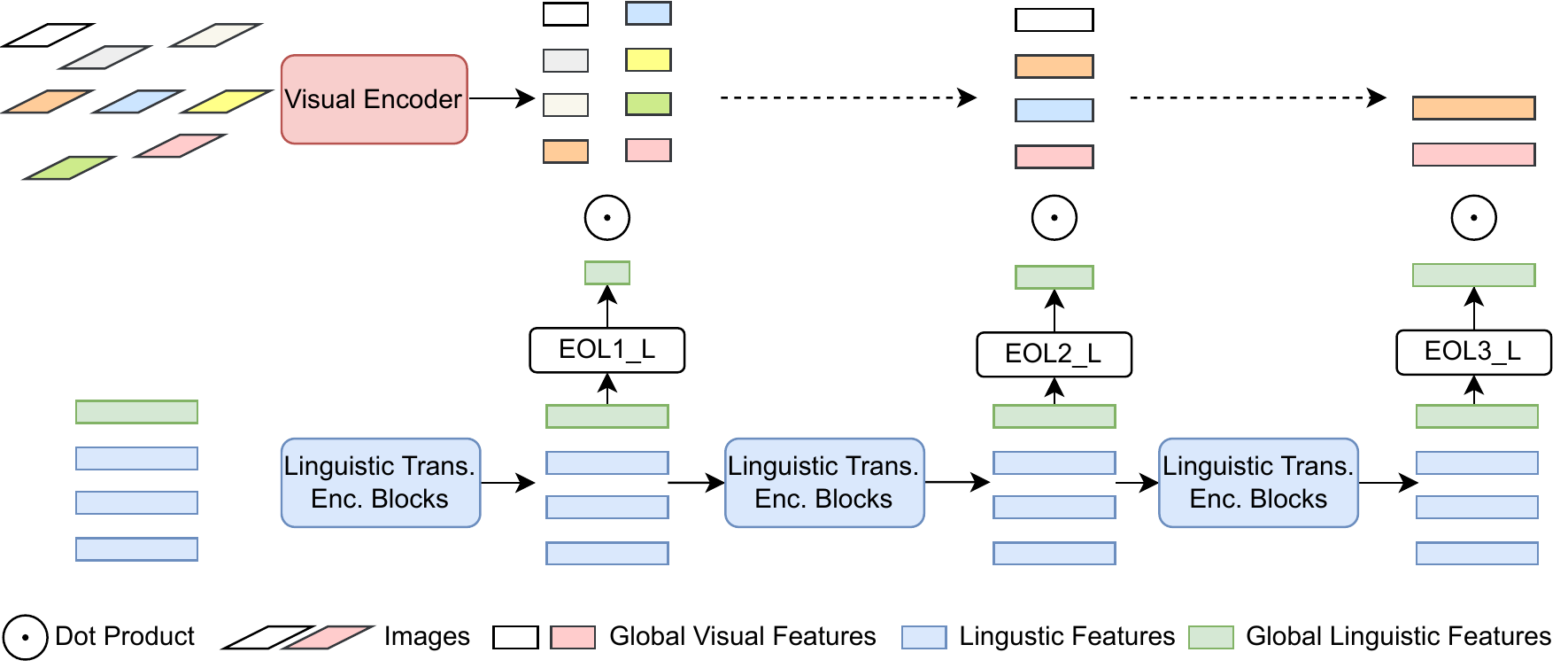}
    \caption{An overview of image retrieval by using our proposed approach HiVLP at inference. Starting from the bottom of the transformer stack, we increase the output dimension of EOL layer-by-layer, to support a coarse-to-fine representation capability, which compute image-text similarity with hierarchical representation to reduce the candidate images and accelerate inference speed.}
    \label{fig:inference}
    \vspace{-3mm}
\end{figure*}

\subsubsection{Hierarchical Retrieval Learning (HRL)}
As shown in Figure~\ref{fig:model}, in order to better carry out image-text retrieval, we leverage paired image-text samples to learn the strong correlation between them at each early output layer. The model is trained to optimize the similarity of paired samples $(I, T)$ and vice versa. The l-th layer similarity of $(I, T)$ is calculated by dot product as follows:
\begin{align}
\label{eq:sim_score}
S^l(t, i) = \hat{\zv}_0^l \odot \hat{\hv}_0^l  ,
\end{align}
where $\odot$ denotes the inner product between two vectors, and $\hat{\hv}_0^l$ and $\hat{\zv}_0^l$ are the output \texttt{[IMG]} and \texttt{[TXT]} representation from $l$-th early ouput layer of visual encoder and linguistic encoder, respectively. We transform $\hv_0^l$ into $\hat{\hv}_0^l \in \mathbb{R}^{d_{EOL_l}}$ via a MLP projection named Early Output Layer (EOL). Similarly, We transform $\zv_0^l$ into $\hat{\zv}_0^l \in \mathbb{R}^{d_{EOL_l}}$ via another MLP projection. In order to utilize training samples efficiently, we propose a in-batch retrieval learning, which close the matched sample and push far the unmatched sample. Given a batch samples with $n$ image-text pairs $(I_i, T_i)$, when we treat text $T_i$ as the query, there are one positive image $I_i$ and  $n-1$ negative images. The text-to-image similarity of $l$-th early output layer is defined as follows:
\begin{align}
    p^l_{t2v, T_i} =  \frac{{\rm exp}({S^l(T_i, I_i)})} { \sum_{k=1}^{n} {\rm exp}({S^l(T_i, I_k)})},
\end{align}
Similarly, we take image $I_i$ as query and compute image-to-text similarity $p^l_{v2t, I_i}$.
The final retrieval loss at $l$-th layer is:
\begin{align}
\label{eq:itr}
    \mathcal{L}^l_{\text{RL}} = \frac{1}{2}\mathbb{E}_{(I,T)\sim D}[\mathrm{CE}(y_{t2v}, p^l_{t2v, T_i}) + \mathrm{CE}(y_{v2t}, p^l_{v2t, I_i})],
\end{align}
where $\rm CE(\cdot)$ denotes the cross-entropy, $y_{t2v}$ and $y_{v2t}$ denotes the ground-truth label and $D$ denotes the training set. For each early output layer, there is a retrieval loss $\mathcal{L}^l_{\text{RL}}$ and the final hierarchical retrieval learning loss is defined as:
\begin{align}
\label{eq:hrl}
    \mathcal{L}_{\text{HRL}} = \sum_{l=1}^{L} \mathcal{L}^l_{\text{RL}}.
\end{align}

\subsubsection{Vision-Language Matching (VLM)} 
To further enhance the model's performance, we utilize the cross-modal encoder to fuse the visual representation and linguistic representation. We regard the fused representation of \texttt{[TXT]} from the cross-modal encoder as multimodal representation. We feed the fused multimodal representation of both modalities to an FC layer and a sigmoid function to predict a score between 0 and 1, where 0 indicates the vision and language are mismatched, and 1 indicates the vision and language are matched as follows:
\begin{align}
\label{eq:vlm}
    \mathcal{L}_\mathrm{VLM} = \mathbb{E}_{(I,T)\sim D} \mathrm{CE} (y, p({\bf z}^{L^V},{\bf h}^{L^T})),
\end{align}
where $D$ is the training dataset, $p(\cdot)$ denotes the transformer decoder blocks and $y$ denotes the ground-truth label.



\begin{table*}[t]
\centering
\small
\resizebox{0.98\textwidth}{!}{
\begin{tabular}{lcccccccccccccccc}
\toprule
\multirow{3}{*}{Model} & \multicolumn{7}{c}{COCO Test (5k images)} & \multicolumn{7}{c}{Flickr30K Test (1k images)}  \\ 
\cmidrule(lr){2-8} \cmidrule(lr){9-15}
& \multicolumn{3}{c}{Text Retrieval} & \multicolumn{3}{c}{Image Retrieval} & & \multicolumn{3}{c}{Text Retrieval} & \multicolumn{3}{c}{Image Retrieval} \\
 \cmidrule(lr){2-4} \cmidrule(lr){5-7} \cmidrule(lr){9-11} \cmidrule(lr){12-14}
& R@1 & R@5 & R@10 & R@1 & R@5 & R@10 & AR & R@1 & R@5 & R@10 & R@1 & R@5 & R@10 & AR \\  \midrule
\multicolumn{15}{c}{Fusion-based Model} \\
\midrule
GXN~\cite{gu2018look} & 42.0	& - &	84.7 & 	31.7 & - &		74.6 & - & 		56.8 & - &		89.6 &	41.5	& - &	80.0 & - \\
SCAN\footnotesize{-single}~\cite{lee2018stacked} & 46.4	& 77.4	& 87.2	& 34.4	& 63.7	& 75.7	& 64.1	& 67.9	& 89.0	& 94.4	& 43.9	& 74.2	& 82.8	& 75.4 \\ 
R-SCAN~\cite{lee2019learning} & 45.4 &	77.9 &	87.9 &	36.2 &	65.6 &	76.7 &	65.0 &	66.3	& 90.6	& 96.0	& 51.4	& 77.8	& 84.9	& 77.8
 \\
BFAN & 49.9 & 79.5 & 88.8 &	36.9 & 65.7 & 77.2	& 66.3	& 69.2 & 91.4 & 96.2	& 50.0 & 77.2 & 84.8	& 78.1\\
CAMP~\cite{wang2019camp} & 50.1 &	82.1	& 89.7	& 39.0	& 68.9	& 80.2	& 68.3	& 68.1	& 89.7	& 95.2	& 51.5	& 77.1	& 85.3	& 77.8 \\
CAAN~\cite{zhang2020context} & 52.5	& 83.3 &	90.9 &	41.2 &	70.3 &	82.9 &	70.2 &	70.1 &	91.6 &	97.2 &	52.8 &	79.0 &	87.9 &	79.8\\
GSMN~\cite{liu2020graph}& 49.4	& 79.3 &	88.8 &	35.9 &	65.5 &	76.9 &	66.0 &	73.3 &	91.8 &	96.4 &	52.4 &	79.2 &	86.3 &	79.9\\
IMRAN~\cite{chen2020imram}& 52.5	& 81.2 &	89.8 &	39.1 &	68.4 &	79.5 &	68.4 &	71.0 &	92.0 &	96.3 &	53.1 &	79.9 &	86.2 &	79.7\\
CAMP-HEI~\cite{tu2021hashing}& 43.4	& 75.0 & 86.2 &	32.3 &	63.2 &	75.1 &	62.5 &	67.6 &	90.9 &	95.0 &	52.6 &	78.2 &	84.4 &	78.1\\
GSMN-HEI~\cite{tu2021hashing}& 49.3&	79.3&	88.8&	35.9&	65.4&	76.7&	65.9
 &	73.4 &	91.9 &	96.7 &	52.4 &	79.0 &	86.0 &	79.9\\
IMRAN-HEI~\cite{tu2021hashing}& 52.5	& 81.3 &	89.8 &	35.9 &	65.4 &	76.7 &	66.9 &	71.0 &	92.0 &	96.4 &	53.2 &	79.7 &	85.8 &	79.7\\
BFAN-HEI~\cite{tu2021hashing}& 49.6	& 79.1 &	88.4 &	36.9 &	65.7 &	77.2 &	66.2 &	69.2 &	91.2 &	96.2 &	49.9 &	77.2 &	84.6 &	78.1\\
SHAN~\cite{ji2021step}& -	& - &	- &	- &	- &	- &	- &	74.6 &	93.5 &	96.9 &	55.3 &	81.3 &	88.4 &	81.7\\
UNITER~\cite{chen2020uniter} & 64.4	& 87.4	& 93.1	& 50.3	& 78.5	& 87.2	& 76.8	& 85.9	& 97.1	& 98.8	& 72.5	& 92.3	& 95.9	& 90.4 \\
OSCAR~\cite{li2020oscar} & 70.0 & 91.1 & 95.5 & 54.0 & 80.8 & 88.5 & 79.8 & - & - & - & - & - & - & - \\
ALBEF \cite{li2021align}   & 73.1          & 91.4          & 96.0         & 56.8          & 81.5          & 89.2 & 81.3  & 94.3         & 99.4          & 99.8       & 82.8          & 96.7          & 98.4  & 95.2\\
VinVL~\cite{zhang2021vinvl} & 74.6 & 92.6 & 96.3 & 58.1 & 83.2 & 90.1 & 82.5 & - & - & - & - & - & - & - \\
ViLT~\cite{kim2021vilt}& 61.5 & 86.3 & 92.7 & 42.7 & 72.9 & 83.1 & 73.2 & 83.5 & 96.7 & 98.6 &64.4 & 88.7 & 93.8& 87.6 \\
Pixel-BERT~\cite{huang2020pixel}& 63.6 & 87.5 & 93.6 & 50.1 & 77.6 & 86.2 & 76.4 & 87.0 & 98.9 & 99.5 & 71.5 & 92.1 & 95.8 & 90.8 \\
Visual Parsing~\cite{xue2021probing} & - & - & - & - & - & -  & - & 87.0 & 98.4 & 99.5 & 73.5 & 93.1 & 96.4  & 91.3 \\
METER~\cite{dou2021empirical}& 76.2 &  93.2 &  96.8 &  57.1 & 82.7 & 90.1 & 82.6 & 94.3 &  99.6 &  99.9 &  82.2 &  96.3 &  98.4& 95.1 \\
\midrule
\multicolumn{15}{c}{Embedding-based Model} \\
\midrule
VSE++~\cite{faghri2017vse++} & 41.3 &	69.2 &	81.2 &	30.3 &	59.1 &	72.4 &	58.9 &	52.9 &	80.5 &	87.2 &	39.6 &	70.1 &	79.5 &	68.3 \\
SCO~\cite{huang2018learning} & 42.8 &	72.3 &	83.0 &	33.1 &	62.9 &	75.5 &	61.6 &	55.5 &	82.0 &	89.3 &	41.1 &	70.5 &	81.1 &	69.9\\
LD~\cite{sun2021lightningdot} & 60.1	& 85.1 &	91.8 &	45.8 &	74.6	& 83.8	& 73.5	& 83.9	& 97.2	& 98.6	& 69.9	& 91.1	& 95.2	& 89.3 \\
LD+OSCAR~\cite{sun2021lightningdot}& 74.2& 92.4 &96.0 &57.4& 82.7 &89.9& 82.1& - & - & - & - & - & -  & - \\
ALIGN$^*$ \cite{jia2021scaling} & \bf{77.0}            & 93.5          & 96.9         & 59.9          & 83.3          & 89.8  & 83.4 & 95.3          & 99.8          & 100.0          & 84.9          & 97.4          & 98.6  & 96.0 \\
\midrule
HiVLP (Ours) & 71.7	& 91.4 &	96.0 &	52.3 &	78.8	& 86.8	& 79.4	& 92.6	& 99.3	& 99.9	& 79.8	& 95.3	& 97.7	& 94.1 \\
HiVLP + CM (Ours) & 76.6	& \bf{94.2} &	\bf{97.1} &	\bf{60.0} &	\bf{83.8} 	& \bf{90.4}	& \bf{83.7}	& \bf{95.5}	& 99.7	& 100.0	& 84.0	& 96.9	& 98.4 & 95.8 \\
\bottomrule
\end{tabular}}
\caption{
\label{tab:retrival_traditional}
Comparision with both fusion-based models and embedding-based models. Our approach achieves comparable performance with a faster inference speed. ``LD'' denotes ``LinghtDot''. * denotes that ALIGN utilizes 1.8B image-text data for training.
}
\end{table*}

\subsection{Real-Time Inference}
As shown in Figure~\ref{fig:inference}, similar to most retrieval systems, our approach consists of two phrases at real-time inference: (1) Offline Representation Extraction and (2) Online Hierarchical Retrieval. We take text-to-image retrieval as an example to introduce our hierarchical retrieval. The purpose of Offline Representation Extraction is to extract the features of candidate images or candidate texts extracted by visual encoder or text encoder offline before retrieval.


\subsubsection{Online Hierarchical Retrieval}
During inference, given a text query $T$, we encode it with our linguistic transformer encoder to obtain hierarchical global representation and compute its similarity with small latency to reduce the candidate images as shown in Figure~\ref{fig:inference}. For $l$-th early output layer, we propose top-$K_l$ retrieval to reduce image candidates and obtain a list of $K_l$ images $I_T$.
At $l+1$-th early output layer, the model only needs to retrieve top-$K_{l+1}$ images from $K_l$ images instead of all the image candidates $K_{ALL}$ ($K_{l+1} \ll K_l \ll K_{ALL}$), which significantly reduces the retrieval time. In addition, the query representation extraction and the similarity computing are processed in parallel approximatively, which further reduces the retrieval time.
\subsubsection{Retrieval Time Analysis}
The retrieval time consumption of embedding-based models comes from the query encoding and the similarity calculation. We assume that the encoding time of each layer is $t_e$, and the calculated time of each dot product similarity is $t_s^d$ for $d$-dimensional vectors. For a text-to-image retrieval, given a query text $T$ with $N$ image candidates, the traditional inference time based on 12 layer encoders is $12t_e+Nt_s^d$. Considering our hierarchical retrieval encoders with 3 early output layers, the inference time of our approach is $4t_e + max(4t_e, N_1t_s^{d_1}) + max(4t_e, N_2t_s^{d_2}) + N_3t_s^d$. When there are $N=10^9$ image candidates and we adopt $N_1=10^9, N_2=10^5, N_3=100$ and $d=768, d_1=128, d_2=300$, under the assumption that each multiplication consumes unit time $1$ and the encoding time $t_e=10^3$, the traditional inference time is about $7 \times 10^{11}$ and our inference time is about $1 \times 10^{11}$. Our proposed hierarchical retrieval is $\sim 7\times$ faster than the traditional embedding-based model settings based on the same model structure.

\section{Experiments}
\subsection{Datasets and Implement Details}
\subsubsection{Datasets}
Similar to most vision-language pre-training models~\cite{chen2020uniter,li2019unicoder,sun2021lightningdot}, we utilize four image-text pairs datasets with 4.1 million images and 9.5 million associated captions for training, including COCO~\cite{lin2014microsoft}, Visual Genome~\cite{krishna2017visual}, Conceptual Captions~\cite{sharma2018conceptual}, and SBU captions~\cite{ordonez2011im2text}. We use Flickr30k~\cite{plummer2015flickr30k} and COCO~\cite{lin2014microsoft} datasets for evaluation, which include 31K images from Flickr30K (29K/1k/1k images for train, validation and test, respectively) and 123K images from COCO (114K/5K/5K for train, validation and test, respectively). Each image from Flickr30k and COCO is associated with 5 human-written captions.

\subsubsection{Implement Details}
We use the visual transformer encoder and linguistic transformer encoder, which has 12/6 layers, 768 hidden states and 12 heads, respectively. The cross-modal encoder has 6 layers, 768 hidden states and 12 heads. We utilize AdamW~\cite{loshchilov2017decoupled} with $\beta_1{=}0.9$, $\beta_2{=}0.98$ to optimize the model training. Moreover, a learning rate warm-up strategy is adopted, where the learning rate 1e-4 is linearly increased during the first 10\% of training steps, followed by a linear decay to 1e-5 in pre-training and 1e-5 to 1e-6 in finetune. The L2 weight decay is 0.01. We set the batch size to 32 per GPU. We conduct pre-training experiments on 8$\times$ A100 GPUs with 6-step gradient accumulation.  We train our model for 10 epochs on Flickr30k, and 10 epochs on COCO.  In our experiments, we utilize 3 EOL layers. We set $N_2=1000$ and $N_3=100$ for all the evaluate sets except for COCO FULL where we set $N_2=5000$ and $N_3=1000$. Notice that $N_1$ equals the number of candidates $N$ of each evaluation set.

\subsection{Evaluation Metrics}
We measure image-text retrieval by recall@$K$ (R@$K$) and the average R@$K$ (AR) for all $K$ for both image-to-text retrieval and text-to-image retrieval tasks. All metrics are the higher the better.

\begin{table*}[t]
\centering

\resizebox{0.97\textwidth}{!}{
\begin{tabular}{lcccccccccccccccc}
\toprule
\multirow{3}{*}{Model} & \multicolumn{7}{c}{COCO-Full (123K Images)} & \multicolumn{7}{c}{Flickr30K-Full (31K Images)}  \\ 
\cmidrule(lr){2-8} \cmidrule(lr){9-15}
& \multicolumn{3}{c}{Text Retrieval} & \multicolumn{3}{c}{Image Retrieval} & & \multicolumn{3}{c}{Text Retrieval} & \multicolumn{3}{c}{Image Retrieval} \\
 \cmidrule(lr){2-4} \cmidrule(lr){5-7} \cmidrule(lr){9-11} \cmidrule(lr){12-14}
& R@5 & R@10 & R@20 & R@5 & R@10 & R@20 & AR & R@5 & R@10 & R@20 & R@5 & R@10 & R@20 & AR \\  \midrule
LightingDot &  40.1 & 51.0 & 62.0 & 28.2 & 37.4 & 47.8 & 44.4 & 69.6 & 78.9 & 86.1 & 51.8 & 62.3 & 72.3 & 70.2 \\ 
HiVLP (Ours) &  54.3 & 65.6 & 69.5 & 42.8 & 55.4 & 60.6 & 58.3 & 78.5 & 94.1 & 97.4 & 60.5 &   84.2 & 90.7 & 84.2\\ 
\bottomrule
\end{tabular}
}
\caption{\label{tab:retrival_full}
Results on the large retrieval setting of full Flickr30k and full COCO datasets.}
\end{table*}

\subsection{Results}
We compare our approach HiVLP with the mainstream state-of-the-art models under the standard retrieval settings and the large candidate settings. The inference speed comparision can be seen in Table \ref{tab:time_cost} for details.

\subsubsection{Base Models}
The compared previous models are divided into two kinds. Fusion-based models: GXN~\cite{gu2018look}, SCAN~\cite{lee2018stacked}, R-SCAN~\cite{lee2019learning}, CAMP~\cite{wang2019camp}, CAAN~\cite{zhang2020context}, ViLBERT~\cite{lu2019vilbert}, Unicoder-VL~\cite{li2019unicoder}, UNITER~\cite{chen2020uniter}, OSCAR~\cite{li2020oscar}, VinVL~\cite{zhang2021vinvl}, ALBEF~\cite{li2021align}, METER~\cite{dou2021empirical} and so on. Embedding-based models: VSE++~\cite{faghri2017vse++}, SCO~\cite{huang2018learning}, LightingDot~\cite{sun2021lightningdot}, ALIGN~\cite{jia2021scaling}. Notice that ALIGN utilizes 1.8B image-text data for training. For fair comparison, we do not compare with ALIGN.

\subsubsection{Results under standard retrieval settings}
Performance on two benchmarks COCO and Flickr30K is shown in Table \ref{tab:retrival_traditional}. ``+OSCAR'' denotes LightingDot uses OSCAR as a Re-Ranker. ``+CM'' denotes we utilize our cross-modal encoder to re-rank the results of the last EOL layer. From the results, we can observe that: 
(1) Compared with embedding-based models. Our approach outperforms non-pre-training approaches (VSE++, SCO) by large margins on all metrics. Specifically, our HiVLP achieves $79.4\%$ on AR on COCO and $94.1\%$ on Flickr30K on AR, beating SCO by $+17.8\%$ on AR on COCO Test and $+24.2\%$ on Flickr30K Test. When comparing with pre-training models, our approach is better than LightingDot on performance, and our HiVLP is much faster than LightingDot (2.23$\times$ faster on Flickr30K-test, 3.33$\times$ faster on COCO-test). Our approach HiVLP achieves better performance compared with the existing fastest pre-training model LightingDot ($+5.9\%$ AR on COCO and $+4.8\%$ AR on Flickr30K) and is still much faster than it (see Table \ref{tab:time_cost}), which shows the efficiency of the proposed hierarchical retrieval structure.

(2) Compared with models with fusion-based. HiVLP also outperforms the non-pre-training models (CXN, CAAN and etc.) by a significant margin. Although  HiVLP achieves slightly lower performance on some metrics than pre-training models with cross-attention, such as the SOTA model METER, HiVLP is much faster than pre-training models with cross-attention. Notice that HiVLP is $\sim$1,400/6,400$\times$ faster than UNITER during inference time and gains 3.7/2.6 points on Flickr30K/COCO simultaneously.

\begin{table}[t]
\centering

\resizebox{0.6\linewidth}{!}{
\begin{tabular}{lccccccccc}
\toprule
\multirow{3}{*}{Model}& \multicolumn{3}{c}{Text Retrieval} & \multicolumn{3}{c}{Image Retrieval} \\
 \cmidrule(lr){2-4} \cmidrule(lr){5-7} 
  & \small{R@1} & \small{R@5} & \small{R@10} & \small{R@1} & \small{R@5} & \small{R@10} & \small{AR} \\  \midrule
HiVLP w/ HRL1 & 56.3 &	81.1 &	87.3	& 36.9	& 67.0	& 78.1	& 66.8 \\
HiVLP w/ HRL2 & 71.9 &	92.4 &	96.8 &	56.2 &	81.0 &	88.7 &	81.2 \\
HiVLP w/ HRL3 & 93.7 &	99.1 &	99.7 &	79.5 &	95.0 &	97.3 &	94.1\\
HiVLP & 92.6	& 99.3	& 99.9	& 79.8	& 95.3	& 97.7	& 94.1 \\
HiVLP + CM & 95.5	& 99.7	& 100.0	& 84.0	& 96.9	& 98.4 & 95.8\\
\bottomrule
\end{tabular}}

\caption{
\label{tab:ablation}
\small{Ablation studies on training strategy over Flickr30K. ``HiVLP w/ HRL1'' denotes we evaluate the model by utilizing the outputs of first EOL layer. ``HiVLP w/ HRL2'' denotes we evaluate the model by utilizing the outputs of second EOL layer. ``HiVLP w/ HRL2'' denotes we evaluate the model by utilizing the outputs of last EOL layer.
}}

\end{table}

\subsubsection{Results under large candidate settings}
To further demonstrate the efficiency of our approach HiVLP, we compare our approach with previous work under large candidate settings following~\cite{sun2021lightningdot} as shown in Table \ref{tab:retrival_full}. HiVLP outperforms LignthingDot with a significant margin ($+14.0\%$ AR on Flickr30K and $+13.9\%$ AR on COCO) with much faster inference speed (5.47$\times$ faster on Flickr30K-Full, 5.05$\times$ faster on COCO-Full as shown in Table~\ref{tab:time_cost}).

\begin{table}[t]
\centering

\resizebox{0.8\linewidth}{!}{
\begin{tabular}{lcccc}
\toprule
Method  &  \#images     & SCAN   & LightingDot & HiVLP (Ours) \\ 
\hline
Flickr30K-test  & 1,000     & 1.8$\times$ &  639$\times$ & 1,427$\times$ \\
COCO-test       & 5,000      & 1.9$\times$  &  1,927$\times$  & 6,426$\times$ \\
Flickr30K-full  & 31,014    & 1.8$\times$  &  6,591$\times$  & 36,051$\times$\\
COCO-full       & 123,287  & 1.9$\times$ &  23,869$\times$ & 120,649$\times$\\
\bottomrule
\end{tabular}}

\caption{Speedup w.r.t. UNITER-base. We compare HiVLP (Ours) and Lightningdot, plus a lightweight cross-attention method SCAN~\cite{lee2018stacked}.}
\label{tab:time_cost}
\vspace{-3mm}
\end{table}

\subsection{Retrieval Time Comparison}
Following LightingDot~\cite{sun2021lightningdot}, we use UNITER as the baseline to compare inference speed in order to demonstrate the efficiency of our HiVLP as shown in Table \ref{tab:time_cost}. In addition, SCAN~\cite{lee2018stacked} is also compared, which uses GRU~\cite{chung2014empirical} instead of a 12-layer Transformer with a more lightweight cross-attention. We test HiVLP and LightingDot on the same GPU and other results are from~\cite{sun2021lightningdot}. As shown in Table \ref{tab:time_cost}, SCAN is $\sim$1.9$\times$ faster than UNITER across both benchmarks because the computational cost of GRU is much cheaper than that of Transformer, but SCAN drops much performance. However, SCAN still utilizes cross-attention and the speedup from SCAN is limited. In contrast, the embedding-based models without cross-attention are much faster than UNITER. Lightningdot is 639$\times$ faster than UNITER on Flickr30K while our HiVLP is 1,427$\times$ faster. HiVLP is still 2.23$\times$ than Lightningdot. When tested with five times more images in COCO, the speedup from HiVLP is 6426$\times$ faster than UNITER while LightingDot is 1927$\times$. HiVLP is 3.33$\times$ faster than Lightningdot.

Our purpose is to build a real-time image-text retrieval system. Thus we test HiVLP in a simulated real-life scenario as shown in Table \ref{tab:retrival_full}. The simulated real-life scenario contains hundreds of thousands of images (text) where we combine all images (text) from training, validation and test set to form a larger candidate pool following~\cite{sun2021lightningdot}. We regard this setting on both benchmarks as Flickr30k-full (31k) and COCO-full (123k), and models are still trained on the training set following~\cite{sun2021lightningdot}. The number of candidate images (texts) scales up by $>$20$\times$ in the simulated real-life scenario. With the same number of text quires, the fusion-based models immediately become impractical. As shown in Table \ref{tab:time_cost}, our HiVLP is 36,051$\times$ faster on Flickr30k-full and 120,649$\times$ faster on COCO-full than UNITER. Moreover, HiVLP is 5.47$\times$ and 5.05$\times$ faster than LightingDot on Flickr30k-Full and on COCO-Full, respectively.

\begin{figure*}[t]
\centering
{\includegraphics[width=0.99\linewidth]{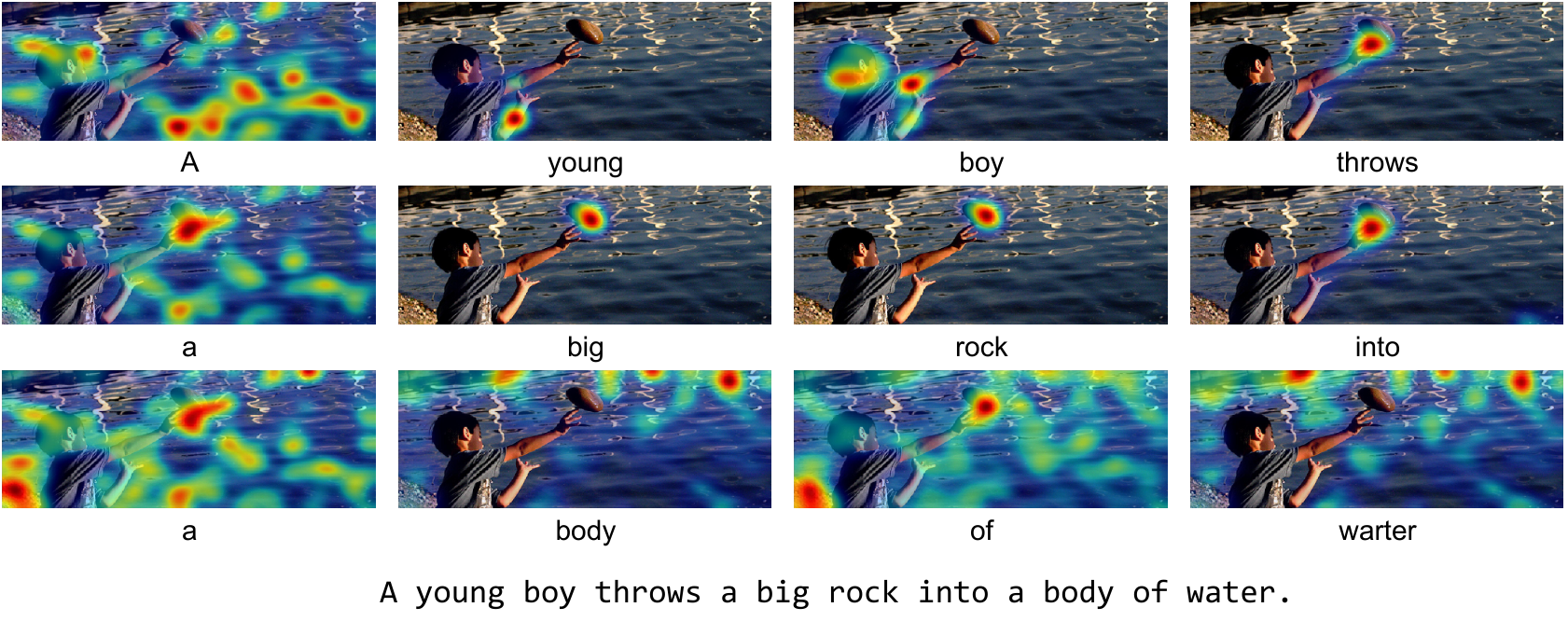}}
\vspace{-6pt}
\caption{Visualization of GradCam for each word. Each image corresponds to a word.}
\label{fig:example1}

\end{figure*}

\subsection{Ablation Study}
In order to comprehensively analyze our model, we conduct the ablation study from the aspect of hierarchical retrieval strategy.

To verify the effectiveness of our hierarchical retrieval strategy, the ablation study is shown in Table \ref{tab:ablation}. We utilize the outputs of different EOL layers to evaluate the models for image-to-text retrieval and text-to-image retrieval on Flickr30K. From the results

\subsection{Qualitative Examples}
\begin{figure*}[t]
\centering
{\includegraphics[width=0.95\linewidth]{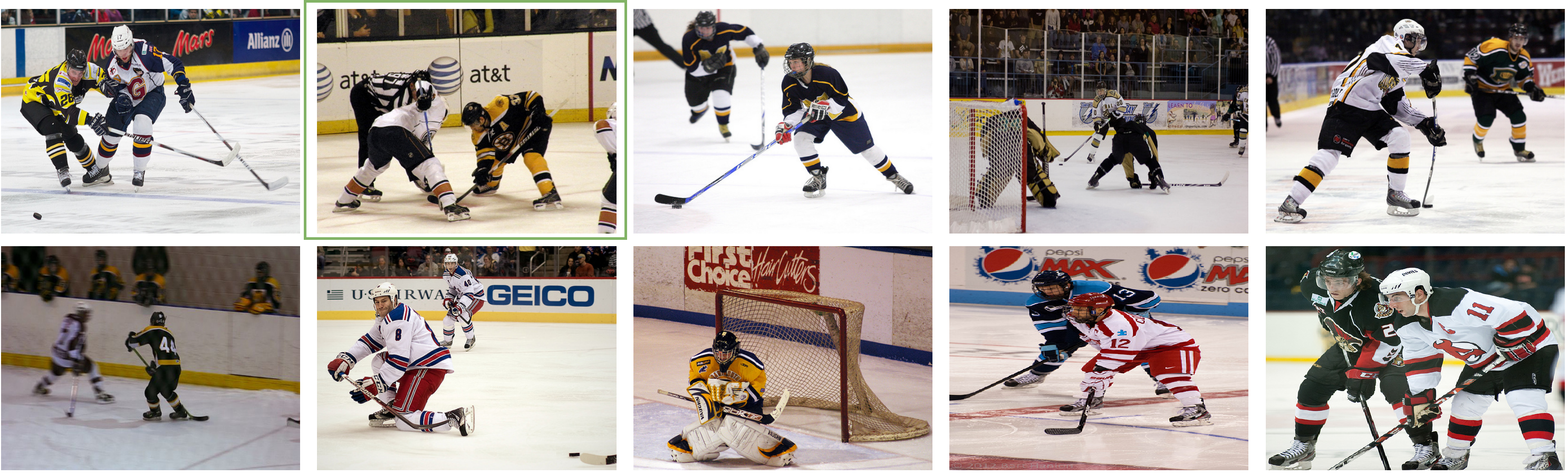}}
\vspace{-6pt}
\caption{Retrieved top 10 images for the query "{\em ice hockey player during the first period of a hockey game against sports team}". The ground truth is in the green rectangle.}
\label{fig:example2}

\end{figure*}

\begin{figure*}[t]
\centering
{\includegraphics[width=0.95\linewidth]{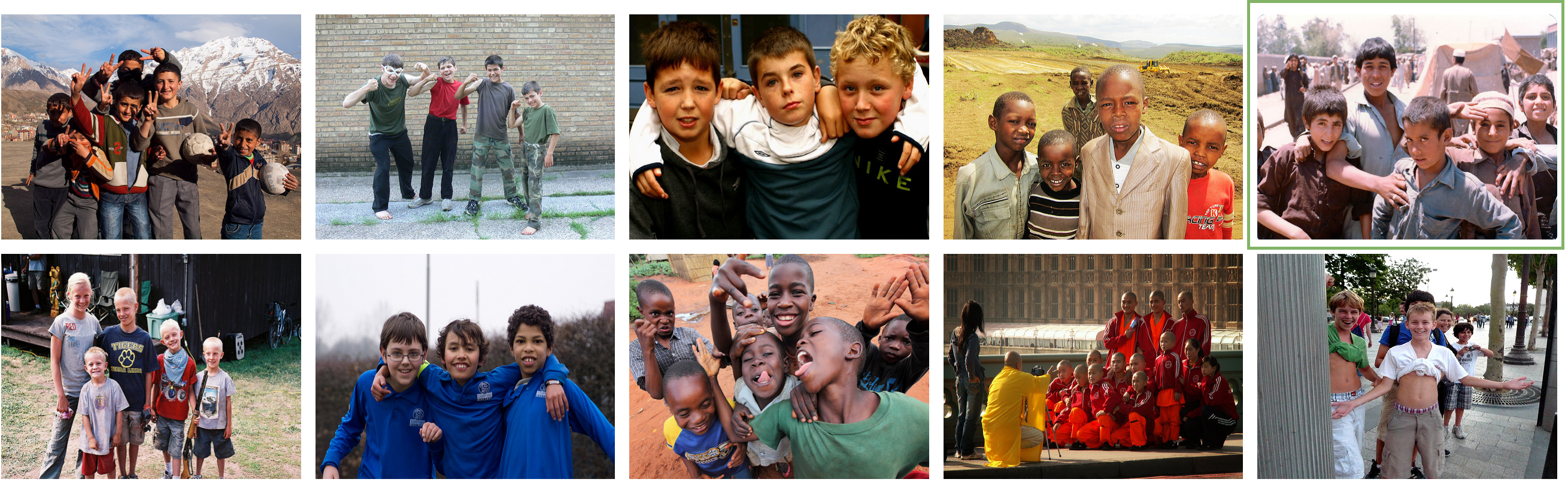}}
\vspace{-6pt}
\caption{Retrieved top 10 images for the query "{\em a group of boys in a crowd}". The ground truth is in the green rectangle.}
\label{fig:example3}

\end{figure*}

\begin{figure*}[t]
\centering
{\includegraphics[width=0.95\linewidth]{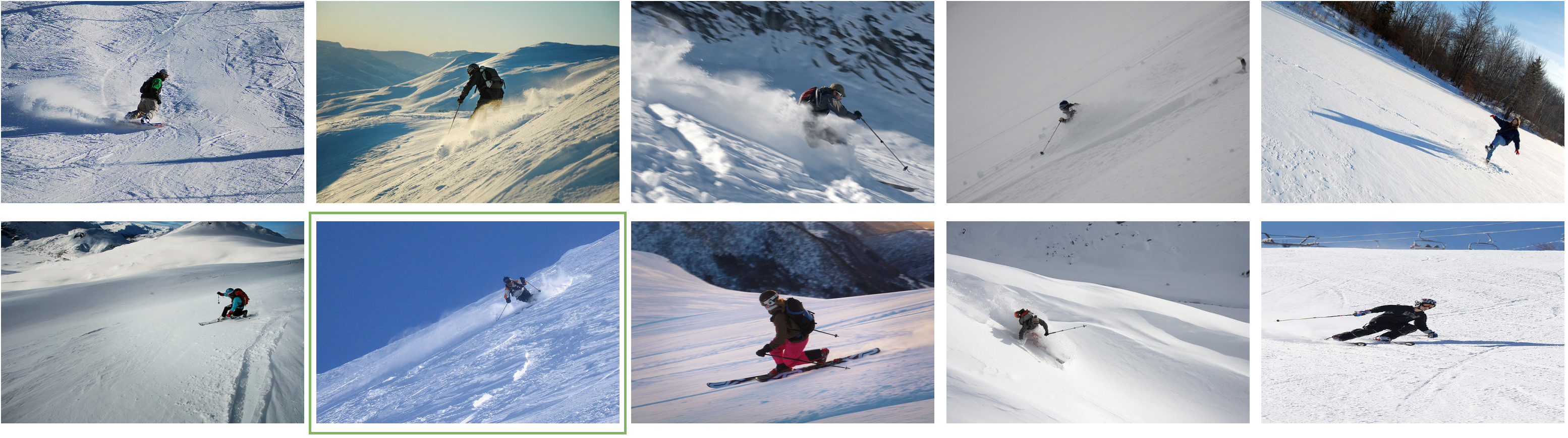}}
\vspace{-6pt}
\caption{Retrieved top 10 images for the query "{\em person on the summit in the snow}". The ground truth is in the green rectangle.}
\label{fig:example4}

\end{figure*}

\begin{figure*}[t]
\centering
{\includegraphics[width=0.95\linewidth]{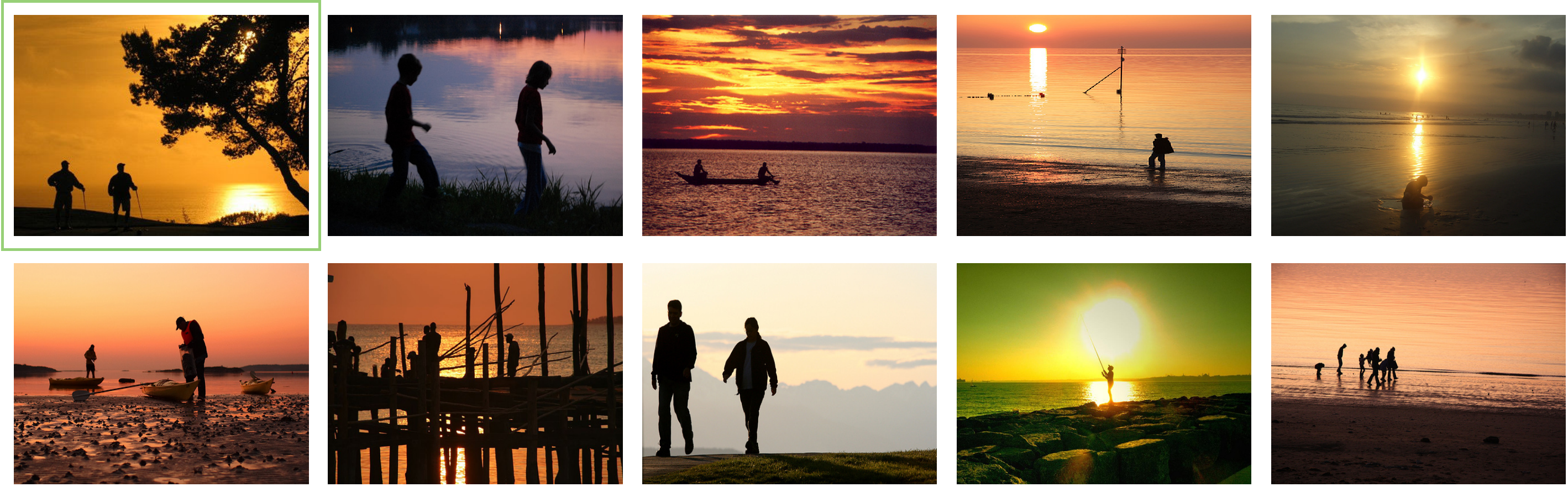}}
\vspace{-6pt}
\caption{Retrieved top 10 images for the query "{\em silhouette of people walking on the beach at sunset}". The ground truth is in the green rectangle.}
\label{fig:example5}

\end{figure*}

As shown in Figure \ref{fig:example1}, we visualize GradCam~\cite{selvaraju2017grad} for each word. The results show that our model aligns words well with visual semantics and thus obtain satisfactory retrieval performance. As shown in Figure \ref{fig:example2}, \ref{fig:example3}, \ref{fig:example4} and \ref{fig:example5}, we show examples of image retrieval results.  All the 10 images are retrieved by our model. The ground truth image is grounded in the green rectangle. Our model usually retrieves the right images in the top 10 candidates, which is satisfies people's sense of experience in multimodal Internet search.

\section{Conclusion}
In this paper, we propose \textbf{Hi}erarchical \textbf{V}ision-\textbf{L}anguage \textbf{P}re-Training (HiVLP) for fast image-text retrieval (ITR). Specifically, we propose a novel hierarchical retrieval objective, which uses the representation of different dimensions for image-text retrieval from coarse-grained to fine-grained retrieval. Specifically, HiVLP use low dimensional representation for large-scale rough retrieval and high-dimensional representation for small-scale fine retrieval. Experiments on two benchmarks show that HiVLP is $1,427$$\sim$$120,649\times$ faster than the fusion-based models UNITER and $2$$\sim$$5\times$ faster than the fastest embedding-based model LightingDot. Moreover, HiVLP achieves the higher performance (about +4.9 AR on COCO and +3.8 AR on Flickr30K than LightingDot) simultaneously.


\clearpage
%
%
\bibliographystyle{splncs04}
\bibliography{egbib}
\end{document}